\documentclass{article}

\usepackage[preprint]{neurips_2025}

\usepackage[utf8]{inputenc} 
\usepackage[T1]{fontenc}    
\usepackage{hyperref}       
\usepackage{url}            
\usepackage{booktabs}       
\usepackage{amsfonts}       
\usepackage{nicefrac}       
\usepackage{microtype}      
\usepackage{xcolor}         
\usepackage{cellspace}
\usepackage{tabularx}

\usepackage{amsmath}    
\usepackage{amssymb}    
\usepackage{amsfonts}   
\usepackage{mathtools}  
\usepackage{bm}         
\usepackage{algorithmic}
\usepackage{algorithm}
\usepackage{verbatim}
\usepackage{authblk}
\usepackage{amsthm}

\newcommand{\overbar}[1]{\mkern 3mu\overline{\mkern-3mu#1\mkern-3mu}\mkern 3mu}
\newcommand{\mt}[0]{\mathbf{\overbar{M}}}
\newcommand{\imt}[0]{\mathbf{\overbar{M}^{-1}}}

\title{Walking the Weight Manifold: a Topological Approach to Conditioning Inspired by Neuromodulation}

\begin{document}

\author[1, *]{Ari S. Benjamin}
\author[1,*]{Kyle Daruwalla}
\author[1,*]{Christian Pehle}
\author[2]{Abdul-Malik Zekri}
\author[1]{Anthony Zador}

\affil[1]{Cold Spring Harbor Laboratory, Cold Spring Harbor, NY}
\affil[2]{University of Southern Florida, Tampa, FL}
\affil[*]{Denotes equal contribution}

\maketitle

\begin{abstract}

One frequently wishes to learn a range of similar tasks as efficiently as possible, re-using knowledge across tasks. In artificial neural networks, this is typically accomplished by \textit{conditioning} a network upon task context by injecting context as input. Brains have a different strategy: the parameters themselves are \textit{modulated} as a function of various neuromodulators such as serotonin. Here, we take inspiration from neuromodulation and propose to learn weights which are smoothly parameterized functions of task context variables. Rather than optimize a weight vector, i.e. a single point in weight space, we optimize a smooth manifold in weight space with a predefined topology. To accomplish this, we derive a formal treatment of optimization of manifolds as the minimization of a loss functional subject to a constraint on volumetric movement, analogous to gradient descent. During inference, conditioning selects a single point on this manifold which serves as the effective weight matrix for a particular sub-task. This strategy for conditioning has two main advantages. First, the topology of the manifold (whether a line, circle, or torus) is a convenient lever for inductive biases about the relationship between tasks. Second, learning in one state smoothly affects the entire manifold, encouraging generalization across states. To verify this, we train manifolds with several topologies, including straight lines in weight space (for conditioning on e.g. noise level in input data) and ellipses (for rotated images). Despite their simplicity, these parameterizations outperform conditioning identical networks by input concatenation and better generalize to out-of-distribution samples. These results suggest that modulating weights over low-dimensional manifolds offers a principled and effective alternative to traditional conditioning.

\end{abstract}

\section{Introduction}

\textit{Conditioning} is essential in the neural network toolbox. For example, an image generator might be conditioned on which category of image to generate or with which style. A running robot might be conditioned on a desired running speed or its goal location. In each of these examples, the sub-tasks are so closely related---relative to all possible functions---that it is more efficient to use a single neural network, provided that one can ensure that learning in one context appropriately generalizes to learning in other contexts.

One way to encourage cross-task generalization would be to take advantage of the clear relationships between these tasks. Here, we pay particular attention to the \textit{topology} of sub-tasks. Running speed is a 1D line through the task space of ambulating. Navigation on a map, conditioned on location, is a 2D operation over a sheet. Classifying images of 3D objects taken from arbitrary orientations lies on the surface of a sphere in task space. An inductive bias which captures these topologies would ensure that knowledge transfers between tasks while enforcing that the true task topology constrains the possible learned input/output mappings.

\begin{figure}[!ht]
    \centering
    \includegraphics[width=0.99\textwidth]{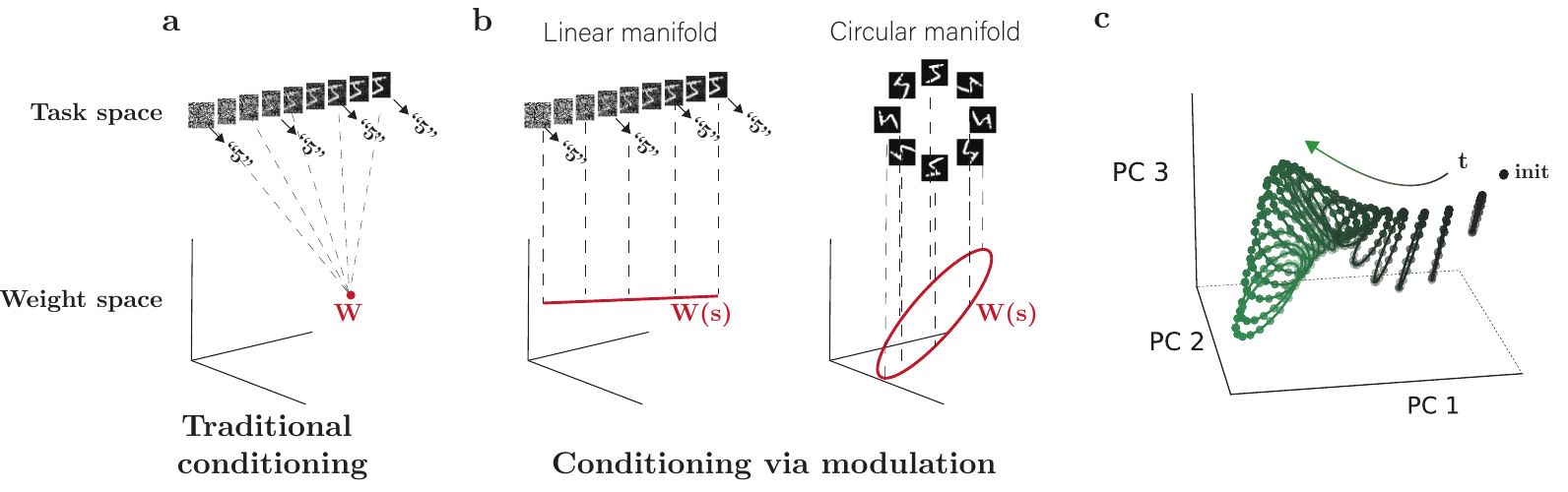}
    \caption{\textbf{a.} Traditional approaches map all conditions in task space to a single network in weight space. Here, conditioning corresponds to the noise added to an image (i.e. input uncertainty) in a classification task. \textbf{b.} Our approach maps various conditions to a parameterized manifold in weight space with known topology chosen to match the task topology. For the task in \textbf{a}, this corresponds to a line, but for alternative tasks like rotations of an input image, an ellipse is more appropriate. \textbf{c.} Manifolds are optimized via our proposed steepest descent rule such that they minimize the volumetric distance between steps. Here, we illustrate this by projecting an ellipse manifold of weights into principal component space during learning. The weights correspond to the first convolutional layer of a network trained to classify CIFAR-10 images.}
    \label{fig:schema}
\end{figure}

Here, we realize this strategy by hypothesizing that for each manifold of tasks there exists a manifold of corresponding topology in weight space which implements that range of tasks. For example, a topological sphere in weight space likely implements the set of functions for classifying objects rotated in 3D space. More formally, assuming smoothness over the weight and task manifolds, and a topology over the task manifold, our ``manifold hypothesis'' postulates a homeomorphism between both manifolds.
Fig.~\ref{fig:schema}a and b illustrate this hypothesis and how it differs from a traditional view on neural network weights and conditioning.

To empirically evaluate this hypothesis, we formalize this notion and derive a learning rule for optimizing manifolds. This rule is analogous to gradient descent, whose implicit learning biases are arguably the key to success in high-dimensional optimization. Just as gradient descent effectively minimizes the path length through weight space, our rule pushes a manifold in the direction of steepest descent while minimizing its total volumetric movement through weight space. This ensures that learning pressure at one location of the manifold does not push other locations on the manifold in very far directions. This learning rule is designed to respect Occam's razor, minimally changing the function across the manifold in response to new data anywhere.

While it might not be obvious that lines, circles, or other simple shapes exist in weight space that implement realistic functions, we find empirically that such solutions do indeed exist. Fig.~\ref{fig:schema}c demonstrates how the rule manipulates the structure of the manifold while taking minimal volumetric steps through weight space (shown here for an image classification network's first layer weights when trained on CIFAR-10 \citep{krizhevsky_learning_2009}). Furthermore, we demonstrate a potential benefit of conditioning via modulation: generalization to contexts not seen during training. Just as humans must learn while calm but perform while nervous or angry, the modulation of artificial networks allows them to generalize performance to areas in which the involved task slightly differs.

In brief, our major contributions include:
\begin{itemize}
    \item a neuro-inspired formalism for topologically constrained weight manifolds
    \item a corresponding steepest descent rule for updating manifolds in principled manner
    \item computationally tractable instances of our rule for simple manifolds (e.g. a line, ellipse, etc.)
    \item practical implementations of our rule that leverage existing automatic differentiation libraries
    \item experimental evaluations demonstrating when the proposed rule is effective and ineffective; notably, that it generalizes to novel conditions unseen in the training data relative to vanilla gradient descent and traditional conditioning
\end{itemize}

\subsection{Inspiration from neuroscience}

The proposed abstraction of weight manifolds reflects a wide set of studies on the physiological properties of neurons and small circuits. After it was observed in the early 20th century that neurons integrate information from their synaptic inputs---directly leading to the earliest generations of neural network models \citep{mcculloch_logical_1943}---it later become clear that nearly all key parameters of neural systems are, in fact, functions of various neuromodulators \citep{harris-warrick_modulation_1991}. While this work vastly complicated the modeling of small biological circuits, it revealed a general biological capability for reusing neural circuits for different purposes in different behavioral states \citep{marder_neuromodulation_2012}.

The mechanistic effects of neuromodulation are profound and diverse. The excitability of neurons are affected, as well as specific synapses and specific ion channels \citep{klein_serotonin_1982, levitan_modulation_1988}. Effects vary depending upon the cell type in question, sometimes in opposite directions for the same neuromodulator \citep{gerfen_modulation_2011}. Furthermore, the delivery of neuromodulation can be extremely targeted to single neurons (i.e. cotransmission) \citep{nusbaum_small-systems_2002}. Together these endow neuromodulation with the potential for strongly affecting the behavior of a single neural network at all functional scales.

In light of this astounding diversity, we aim to abstract only a very general principal of operation. Notably, we are not attempting to model particular functions of various neuromodulators in different circuits. Instead, we ask what advantage this general capability provides to artificial systems.

\subsection{Modulating a network $\subset$ conditioning a network}
\label{sec:high-v-low-conditioning}

As a note of clarity, conditioning information may be either \textbf{high-dimensional} (for example, CLIP embeddings to generate images from natural language descriptions) or \textbf{low-dimensional} (for example, desired running speed of a robot). Here, we are solely interested in the low-dimensional case, which we believe admits special treatment. To distinguish low-dimensional conditioning cases, we will refer to this as \textit{modulating} a neural network.

Low-dimensional modulations of a network are quite common. A helpful example can be found in generative modeling via diffusion (DDPMs) where the denoiser is conditioned on the magnitude of injected noise, or more specifically, the timestep of the diffusion process \citep{ho_denoising_2020}. This allows a single denoising neural network to denoise both low-noise and high-noise examples, but transfer knowledge between noise regimes. 

\section{Optimizing a weight manifold: a formal treatment}
\label{sec:theory}

Neural networks are typically optimized as a single point in weight space; however, our approach requires optimizing an entire manifold of weights simultaneously. This section formalizes this approach while providing intuition for the underlying concepts.

\subsection{Weights as parameterized functions}

In traditional neural networks, each weight (synapse) is described by a single value. Here, we make each weight a function of the conditioning variable. Formally, consider a smooth manifold $\mathcal{M}(s, \mathbf{P})$ parametrized by $s \in [0,1]$ and depending on parameters $\mathbf{P} \in \mathbb{R}^p$. The parameter $s$ represents the \textbf{modulator} or conditioning value, and the parameters $\mathbf{P}$ are learnable. Then, $\mathcal{M}: \mathbb{R} \times \mathbb{R}^p \to \mathbb{R}^d$ is map from points on the manifold, selected by $s$, to the weights of the network.

\paragraph{Example:} A simple manifold is a straight line segment through weight space. Just as a line can be uniquely defined by its two endpoints, we can parametrize this manifold as a linear interpolation:
\begin{equation}
    \mathcal{M}(s, \mathbf{P}) = (1-s) \mathbf{P}_1 + s \mathbf{P}_2
\end{equation}
where $\mathbf{P} = (\mathbf{P}_1, \mathbf{P}_2)$ contains the endpoints. When $s=0$, we get the weights $\mathbf{P}_1$; when $s=1$, we get $\mathbf{P}_2$; and for values in between, we get a smooth interpolation along the line.

\subsection{Optimization}

Unlike standard neural network training that minimizes a loss at a single weight configuration, we need to minimize a loss \textbf{functional} $L[\mathcal{M}]$ that evaluates the entire manifold:
\begin{equation}
    L[\mathcal{M}] = \int_0^1 \ell(\mathcal{M}(s, \mathbf{P})) \; \mathrm{d}s
\end{equation}
where $\ell: \mathbb{R}^d \to \mathbb{R}$ is a function measuring the loss at each point along the manifold.

\paragraph{Intuition:} We're essentially averaging the performance across all possible conditioning values. This ensures that our network performs well across the entire conditioning spectrum, not just at isolated points.

\subsubsection{The variational problem}

At each optimization step, we need to find the best way to update our parameters $\mathbf{P}$. This means finding the optimal perturbation, $\Delta \mathbf{P}$, that moves the entire manifold in a beneficial direction.

\begin{alignat}{3}
&\Delta \mathbf{P} =&& \arg\min_{\Delta \mathbf{P}} && \: L[\mathcal{M} (s, \mathbf{P} + \Delta \mathbf{P})] \\ \nonumber
&&&\text{such that } && \: d^2(\mathcal{M}(s, \mathbf{P} + \Delta \mathbf{P}), \mathcal{M}(s, \mathbf{P})) = C_0
\label{eq:steep-descent-obj}
\end{alignat}

where $d^2$ denotes the total squared distance over the manifold (i.e. the Euclidean distance at every point integrated over $s$). The distance constraint ensures that the manifold does not move too far in a single step.

\paragraph{Key insight:} Just as gradient descent minimizes distance traveled through weight space, our approach minimizes the \textbf{volumetric movement} of the entire manifold. This ensures that learning at one point on the manifold does not cause excessive changes elsewhere.

\subsubsection{Solving for $\Delta \mathbf{P}$} 

First, we note that small perturbations affect our manifold approximately linearly:
\begin{align}
    \mathcal{M}(s, \mathbf{P} + \Delta \mathbf{P}) \approx \mathcal{M}(s, \mathbf{P}) + \frac{\partial\mathcal{M}}{\partial \mathbf{P}}\Delta \mathbf{P} 
\end{align}

Thus, the perturbation $\Delta \mathbf{P}$ affects the manifold dependent upon the Jacobian of the parameterization, $\frac{\partial\mathcal{M}}{\partial \mathbf{P}}$. To express the optimal update, it is helpful to introduce the following notation:

\begin{align}
    \mathbf{M}(s) &= \left(\frac{\partial\mathcal{M}}{\partial \mathbf{P}}\right)^T\frac{\partial\mathcal{M}}{\partial \mathbf{P}} & \text{(local metric tensor)} \\
    \mathbf{g}(s) &= \left(\frac{\partial\mathcal{M}}{\partial \mathbf{P}}\right)^T\nabla \ell(\mathcal{M}(s)) & \text{(local gradient w.r.t. $\mathbf{P}$)}
\end{align}

The metric tensor $\mathbf{M}(s)$ captures how parameter changes affect the manifold at each point $s$, while $\mathbf{g}(s)$ represents the direction of steepest descent of the loss at each point.

Through Lagrangian optimization on Eq. \ref{eq:steep-descent-obj} (detailed derivation in the supplemental information), we arrive at the optimal parameter update:

\begin{equation} \label{eq:aupdate}
    \Delta \mathbf{P} = -\frac{1}{2\lambda} \left[\int_0^1 \mathbf{M}(s) \; \mathrm{d}s \right]^{-1}\int_0^1 \mathbf{g}(s) \; \mathrm{d}s
\end{equation}

This update has an intuitive interpretation: we're averaging gradients across the entire manifold and then applying a correction based on the manifold's geometry.

In practice, the integral over the gradients $\int_0^1 \mathbf{g}(s) \; \mathrm{d}s$ can be computed via sampling without modifications to stochastic gradient descent frameworks.
The term in brackets, $\left[\int_0^1 \mathbf{M}(s) \; \mathrm{d}s\right]^{-1}$, is the inverse of the integrated metric tensor, which we will denote this as $\imt$. For general manifolds, $\imt$ may be challenging to calculate as it represents the inverse of a very large matrix (with as many rows as network weights). Luckily, this term is analytically computable in several special cases that are relevant for practice, such as lines, ellipses, or any parameterized manifold expressible as a weighted sum of certain basis points.

\subsubsection{Manifolds with analytic $\imt$}
\label{sec:analytic-manifolds}

For many manifold types, we can analytically determine the inverse integrated metric tensor $\imt = \left[\int_0^1 \mathbf{M}(s) \; \mathrm{d}s\right]^{-1}$ and its matrix-vector product. This allows for efficient optimization without needing to compute and invert large matrices.

\paragraph{Example: Straight Line Manifold}

To illustrate this, let's consider again the straight line manifold parametrized as a linear interpolation between two points $\mathcal{M}(s, \mathbf{P}) = (1-s) \mathbf{P}_1 + s \mathbf{P}_2$.

To compute the metric tensor at any point $s$, we need the Jacobian:

\begin{equation}
\frac{\partial \mathcal{M}}{\partial \mathbf{P}} = \begin{bmatrix} (1-s)\mathbf{I} & s\mathbf{I} \end{bmatrix}
\end{equation}

where $I$ is the identity matrix with the same dimension as all network parameters, flattened. The metric tensor is then:

\begin{equation}
\mathbf{M}(s) = \left(\frac{\partial\mathcal{M}}{\partial \mathbf{P}}\right)^T\frac{\partial\mathcal{M}}{\partial \mathbf{P}}=\begin{bmatrix} (1-s)^2 \mathbf{I} & s(1-s) \mathbf{I} \\ s(1-s) \mathbf{I}& s^2 \mathbf{I} \end{bmatrix}
\end{equation}

The inverse of this matrix after integrating over $s$ from 0 to 1 is our desired $\imt$:

\begin{equation}
\imt = \left[\int_0^1 \mathbf{M}(s) \; \mathrm{d}s\right]^{-1} = \hspace{2pt}\begin{bmatrix} \frac{1}{3}\mathbf{I} & \frac{1}{6}\mathbf{I} \\ \frac{1}{6}\mathbf{I} & \frac{1}{3}\mathbf{I} \end{bmatrix}^{-1}= \begin{bmatrix} 4\mathbf{I} & -2\mathbf{I} \\ -2\mathbf{I} & 4\mathbf{I} \end{bmatrix}
\end{equation}

This may seem like too a large matrix to hold in memory; however, it never needs to be instantiated---all that is required to calculate $\Delta P$ is its matrix-vector product with the integrated local gradient. This is simple to evaluate. Plugging into Equation \ref{eq:aupdate}, we obtain:

\begin{equation}
\Delta \mathbf{P} = -\frac{1}{2\lambda}\begin{bmatrix} 
4\nabla_{\mathbf{P}_1} \ell(\mathcal{M}(s)) - 2\nabla_{\mathbf{P}_2} \ell(\mathcal{M}(s)) \\ 
-2\nabla_{\mathbf{P}_1} \ell(\mathcal{M}(s)) + 4\nabla_{\mathbf{P}_2} \ell(\mathcal{M}(s)) 
\end{bmatrix}
\end{equation}

Thus, this method is easy to implement and consists only of fast linear combinations of gradients computed through automatic differentiation. By comparison, if we had not considered the metric penalty, the update rule would simply be the gradient $\begin{bmatrix} \nabla_{\mathbf{P}_1} l(\mathcal{M}(s)) & 
\nabla_{\mathbf{P}_2} l(\mathcal{M}(s))
\end{bmatrix}^T$, without any mixing of the gradients with respect to each endpoint.

\paragraph{Summary of Analytical Cases:}

Many useful manifold parameterizations have closed-form expressions for $\imt$, making them computationally efficient. Table~\ref{tab:manifolds} summarizes key examples.

\begin{table}[!ht]
\centering
\caption{Analytical forms of $\imt$ for common manifold types}
\label{tab:manifolds}
\begin{tabular}[t]{SlSlSl}
\toprule
\textbf{Manifold Type} & \textbf{Parameterization} & $\imt$ \\
\midrule
Straight Line & $(1-s) \mathbf{P}_1 + s \mathbf{P}_2$ & $\begin{bmatrix} 4\mathbf{I} & -2\mathbf{I} \\ -2\mathbf{I} & 4\mathbf{I} \end{bmatrix}$ \\
Ellipse & $\mathbf{P}_1 + \mathbf{P}_2\cos(2\pi s) + \mathbf{P}_3\sin(2\pi s)$ & $\begin{bmatrix} \mathbf{I} & 0 & 0 \\ 0 & 2\mathbf{I} & 0 \\ 0 & 0 & 2\mathbf{I} \end{bmatrix}$ \\
Tethered Rod & $(1-s)\mathbf{P}_1 + s\mathbf{P}_2$ & $\begin{bmatrix} 0 & 0 \\ 0 & 3\mathbf{I} \end{bmatrix}$ \\
Cubic B-Splines & $\sum_{i=0}^{n} N_{i,k}(s) \mathbf{P}_i$ & \parbox{5cm}{$\mathbf{D}^{-1}$ where $\mathbf{D}$ is banded diagonal;\\ $\frac{1}{60n}[120, 78, 24, 3]$ along diagonals} \\
\bottomrule
\end{tabular}
\end{table}

\subsection{Practical Implementation: Efficient Optimization of Weight Manifolds}
\label{sec:manifold-implementation}

During learning and inference, one sees a batch of examples with varying levels of conditioning, $\{(\mathbf{x}_i, s_i)\}_i^B$. The challenge is to obtain the output for each example, noting that each different values of $s$ correspond to the outputs of effectively different neural network with weights $\mathcal{M}(s, \mathbf{P})$. It would be inefficient to instantiate each neural network separately to process each example, as the memory requirements would scale linearly with the batch size, $B$.

The key insight of our implementation is that for all of the manifolds shown in Table~\ref{tab:manifolds}, the weight vector at a specific point on the manifold, $\mathcal{M}(s, \mathbf{P})$, is a \textbf{linear combination of $n$ ``basis points''} $\mathbf{P}_i$. Each basis point defines a particular instance of the network.
\begin{equation}
    \mathcal{M}(s, \mathbf{P})=\sum_{i=0}^{n} a_i(s) \mathbf{P}_i
    \label{eq:linear-manifold-prop}
\end{equation}
Conveniently, most learnable operations in neural networks (e.g. fully-connected layers, convolutions, embeddings, sub-operations of self-attention, etc.) are also linear in that they are additive and homogenous---$f(\alpha a+\beta b)=\alpha f(a) +\beta f(b)$. Nonlinear operations such as ReLU generally do not contain any learnable components. This property along with Eq.~\ref{eq:linear-manifold-prop} allows us to process all $B$ examples in a batch while only ever holding the $n$ basis points in memory.

The reason for this is that within each layer, the effective $s$-dependent weight matrix $\mathbf{W}(s)$ never needs to be instantiated to obtain the matrix-vector product $\mathbf{W}(s) \mathbf{x}$. Instead, one can apply each basis point separately, then linearly combine them: 
\begin{equation}
    \mathbf{W}(s) \mathbf{x} = a_1(s) \mathbf{W}_1 \mathbf{x} + a_2(s) \mathbf{W}_2 \mathbf{x} + \ldots
\end{equation}
Often, $n$ is small enough that each term in the linear combination can be computed in parallel. This procedure can then be extended layer-by-layer to cover an entire neural network. Thus, with minimal memory overhead and no runtime overheads, we can compute a full batch of examples over the manifold despite $B \gg n$.

Algorithm~\ref{alg:manifold_optimization} (in appendix) provides a high-level overview of our approach.

\section{Empirical results}
\label{sec:experiments}

In evaluating our proposed approach, we must consider when our neuromodulation-inspired abstraction is useful for training artificial networks. To this end, we study two settings where topological constraints on network weights are effective but also where they are ineffective. First, Sec.~\ref{sec:generalization} demonstrates the generalization ability of our approach to unseen data augmentation conditions when the augmentation topology is known. Surprisingly, this occurs even for simple, rigid manifolds, experimentally proving our ``manifold hypothesis.'' Second, Sec.~\ref{sec:regularization} attempts to leverage manifolds to regularize networks in the face of uncertainty where the mapping from conditioning value to task manifold is more complex. 
\emph{Our aim in these experiments goes beyond validating the correctness of our approach; we hope to highlight both the use cases and pitfalls of using modulation as a mechanism to address a learning challenge.} Please refer to the appendix for complete details on reproducing these results.

\subsection{Generalization to unseen conditions}
\label{sec:generalization}

\begin{figure}[!ht]
    \centering
    \includegraphics[width=0.99\textwidth]{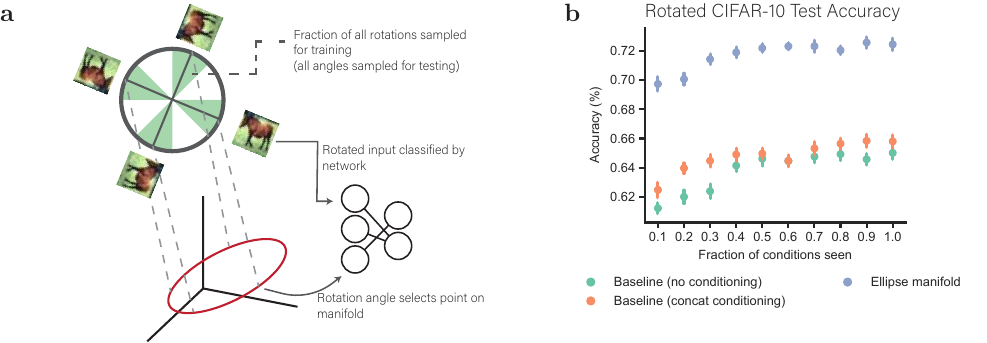}
    \caption{\textbf{a.} An illustration of the training paradigm used to test the generalization abilities of our manifold approach. We sample a sparse subset of the possible conditions (rotation angles of the input) during training and test on the full set of conditions. \textbf{b.} Performance of the ellipse manifold network on the test set vs. a baseline network with and without conditioning.}
    \label{fig:conditioning-sparsity}
\end{figure}

Here, we design a setting where data augmentation is applied the inputs to the network, and the conditioning describes the augmentation that was applied. Specifically, we rotate CIFAR-10 images passed to a moderately sized convolutional network (CNN) and the conditioning, $s$, encodes the angle of the rotation from $[0, 2\pi]$. The network must classify these images despite the rotation. To test generalization, we only sample a fixed subset of the possible angles during training (e.g. 10\% of $[0, 2\pi]$). During evaluation, we sample the full range and report the test accuracy as a function of the sparsity of the training conditions. Fig.~\ref{fig:conditioning-sparsity}a illustrates the overall setting.

As shown in Fig.~\ref{fig:conditioning-sparsity}b, training an ellipse manifold generalizes to unseen angles even at extremely sparse sampling of conditions during training. Furthermore, the performance of the manifold network surpasses conventional networks with and without conditioning input. We also note that the baselines lose performance as the training conditions become sparser, while the manifold networks maintains its peak performance for a wider range of sampling sparsities.

We emphasize that these results are non-trivial. They demonstrate, empirically, that the ``manifold hypothesis'' is true, and that we can match the appropriate homeomorphism between the task and weight manifolds with even a simple parametrized ellipse. The manifold used here is simple in that it is rigid---it can rotate and stretch in high dimensional weight space, but it cannot bend and must lie in a plane. This is an extremely strong constraint on space the weights can occupy while still performing the task correctly and generalizing. Yet, when the topological structure in the data is clear (such as rotation invariance of images), modulating weights with even a simple mechanism is sufficient to exploit this structure.

\subsection{Controllable regularization of networks}
\label{sec:regularization}

While known data transformations such as augmentation have clearly definable topologies that can be exploited, this is not the only form of conditioning that our approach could target. An unknown data transformation that must frequently be dealt with is input noise, and models must robustly perform in the face of this uncertainty. Regularization of a networks weights is a common attempt for dealing with noisy data. Traditionally, a single network is regularized by a fixed amount which may be overly conservative when the noise varies from sample to sample.

Here, we study a setting where a network classifies images from CIFAR-10 with additive Gaussian noise. Specifically, for each example, we sample a noise level $s \sim \text{Unif}(0, S)$ where $S$ denotes the maximum noise level. Then, we augment the input image as $\hat{\mathbf{x}} = (1 - s) \mathbf{x} + s \eta$ where $\eta \sim \mathcal{N}(0, \mathbf{I})$. Thus, $s$ reflects the uncertainty of the example. We train a CNN whose weights are constrained to a line manifold with $L_2$-regularization. For a given sample, $(\mathbf{x}_i, \mathbf{y}_i, s_i)$, the loss of that sample is
\begin{equation}
\ell_\mathcal{M}(\mathbf{x}_i, \mathbf{y}_i, s_i; \mathcal{M}(s_i, \mathbf{P})) = \ell(\mathbf{x}_i, \mathbf{y}_i; \mathcal{M}(s_i, \mathbf{P})) + s_i \lambda \|\mathcal{M}(s_i, \mathbf{P})\|_2^2
\end{equation}
where $\ell$ is the original loss (e.g. cross entropy), $\mathcal{M}(s_i, \mathbf{P})$ are the network weights used for the $i$-th sample, and $\lambda$ is a regularization coefficient. Thus, our uncertainty conditioning controls the level of regularization applied to each example.

\begin{figure}[!ht]
    \centering
    \includegraphics[width=0.99\textwidth]{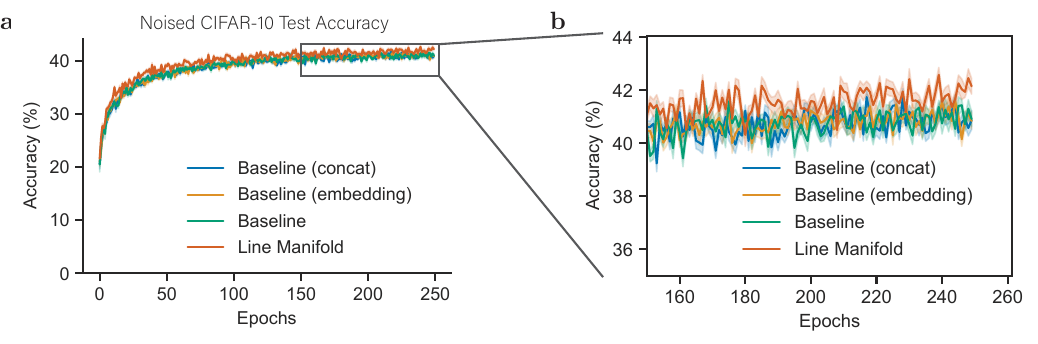}
    \caption{\textbf{a.} Noised CIFAR-10 test accuracies for baseline networks and the line manifold. Noise procedure is described in Sec.~\ref{sec:regularization}. \textbf{b.} A zoomed-in view of \textbf{a}.}
    \label{fig:regularized-cifar}
\end{figure}

Fig.~\ref{fig:regularized-cifar} shows the test performance on CIFAR-10 (which is also noised in the same manner). While the manifold network does slightly outperform the baseline cases, its advantage is minimal. We hypothesize that this is because of the mismatch between our specification of uncertainty and the true uncertainty manifold. Namely, only the few examples closed to the learned decision boundary are relevant to the model's uncertainty, and thus, our chosen manifold spans a much wider range of the task space than the true manifold. These results illustrate cases where modulation is not useful---when the topology is difficult to define or properly infer and relate to the conditioning value.


\section{Related work}

\paragraph{Dynamic weights:} Our work and several other works share the general idea that weight matrices might be adaptive, rather than fixed. For example, fast weight programmers, dynamic filter networks, and linear transformers can all be seen as having weights which are themselves a function of the input \citep{ba_using_2016, jia_dynamic_2016, schlag_linear_2021}. These methods differ from our approach in that the contexts are inferred rather than supplied as a method for external conditioning. Furthermore, these methods do not constrain the movement of the implied manifold over weights, nor ensure that its topology matches the task at hand.

\paragraph{Linear modes and distributions over networks:} Weight manifolds are one way to establish a set of networks that work well. In this way, it is closely related to hypernetworks \citep{ha_hypernetworks_2016} and Bayesian networks \cite{mackay_practical_1992, barber_ensemble_1998}, which both establish distributions over weight space. More close in spirit is work which documents the existence of ``linear mode connectivity,'' i.e. that paths of everywhere-low loss exist in weight space which connect separate learning trajectories \citep{draxler_essentially_2018}. Such paths are also empirically found connecting minima from related tasks, i.e. connecting the pretrained multitask solution with a fine-tuned solution \citep{mirzadeh_linear_2020}. Here, rather than find such solutions empirically, we provide a framework for directly training lines in weight space, and in general, any low-dimensional manifold. Interestingly, our results provide proof that there exist straight lines that connecting modes, not only the smoothly curved lines found \emph{post hoc} after training as in prior works.

\paragraph{Conditioning methods:} Although conditioning via input concatenation and embedding are standard, several other methods for conditioning exist. One closely related work is FiLM, which learns to apply an affine transformation to the network's intermediate activations which is a function of the conditioning variable \citep{perez_film_2017}. Other methods with a similar philosophy include Conditional Instance Norm \citep{dumoulin_learned_2017} and Conditional Batch Norm \citep{de_vries_modulating_2017}, which adapt standard layer normalization layers to be functions of the conditioning information. These methods effectively establish a manifold over the biases of each layer, which is either an affine manifold (in the case of FiLM) or curved according to the divisive normalization scheme. Our approach can be seen as a generalization of these methods.

\paragraph{Models of neuromodulation:} Weight manifolds can be seen as an abstraction and generalization of many computational models of neuromodulation. While no single paragraph can summarize all such models, here we highlight those models in which neuromodulation acts as a functional knob upon circuit behavior \citep{abbott_modulation_1990, brezina_analyzing_1997, stroud_motor_2018, vecoven_introducing_2020, osman_hopfield_2024, costacurta_structured_2024, tsuda_neuromodulators_2024}. For example, one classic model describes the effect of acetylcholine in the hippocampus as a knob upon top-down/bottom-up gain in a generative model, effectively correlating acetylcholine to perceptual uncertainty \citep{yu_acetylcholine_2002, yu_uncertainty_2005}. While details differ, these papers generally supply specific mechanisms in which neuromodulation affects circuit behavior. We argue that each of these models are equivalent to a manifold in weight space of a single network, and furthermore that it is productive to consider the abstract properties of such manifolds for learning and computation.

\section{Broader impact}
\label{sec:impact}

This work is focused on foundational theoretical research for optimizing manifolds of model weights. As such, it does not have direct deployment considerations or immediate negative harms. Still, a plausible positive impact of this work is better control, interpretation, and constraint of learned neural network functions. This should not be misconstrued for safety guarantees---networks learned with our approach are only constrained along the conditioning axes specified by the researcher. Misalignment between the intended axes and the specification can result in unexpected behavior, and more importantly, the model is not constrained on unspecified axes. Finally, an common application of conditioning is in large language models and generative models (e.g. image diffusion models), so the improvements in this work endow these models with additional capabilities, potentially exacerbating existing harms and misuse.

\section{Discussion}

Here, we demonstrated how the general principle behind neuromodulation can be ported to artificial neural networks by analogy to low-dimensional manifolds in weight space. We showed experimentally that simple manifolds (such as straight lines and ellipses) that solve tasks indeed exist in weight space, and furthermore, can be trained using a novel steepest rule for manifolds. Our approach provides a robust and principled procedure to create collectives of networks that learn together yet differ from one another in meaningful ways. 

We conjectured that the advantage of conditioning by modulating weights instead of injecting input would be the ability to exploit the topology of the data. Thus, the choice of manifold provides a programmable inductive bias for conditioning. In support of this claim, we demonstrated that this allows for out-of-distribution generalization to unseen conditioning values better than conditioning via injecting input. On the other hand, when the chosen manifold and task topology are misspecified, modulating weights provides limited advantages. Thus, our experimental results demonstrate that modulating weights is useful computational primitive when the topology of the task is clear, and it is easily exploited by simple modulation schemes (i.e simple manifolds).

Our theoretical formulation provides a foundation for many potential applications beyond those that we studied in this paper. In particular, we focused on a few simple manifold types, but future extensions could include more complex manifolds that can be bent and distorted in weight space to permit more flexible topology embeddings. Alternatively, two, three, or higher dimensional topologies would permit the exploration of how different conditions across tasks interact in weight space. Furthermore, we study settings where the conditioning value is explicit and known, but this is the rarely the case for biological networks. Instead, a more realistic case should explore inferring or controlling the conditioning value through learned experience. Finally, an important use case for conditioning is encouraging a model to be invariant or equivariant to data transformations. Unlike many existing methods, our framework allows a network to target either case and potentially explore the trade-off between the two. Ultimately, the success of deep learning models has been their ability to decipher and exploit structure in the world. While this is typically statistically gleaned from data, inductive biases that are strong yet flexible allow networks to learn more efficiently and generalize. Our work, through a formal treatment of the connection between functions embedded in weight space and topologies in task space, enables a new generation of programmable, flexible, and controllable inductive biases for neural networks.



\section*{Code availability}

The code for all figures in this paper were written in Jax \citep{bradbury_jax_2018} and will be made available shortly.

\bibliography{references}
\bibliographystyle{plain}

\newpage
\appendix

\section{Supplement: Mathematical details of manifold optimization}
\label{apx:theory}

\subsection{Theorem 1: Optimal Manifold Perturbation}
\textbf{Theorem 1.} \textit{Given a parametrized manifold $\mathcal{M}(s, \mathbf{P})$ with loss functional $L[\mathcal{M}] = \int_0^1 \ell(\mathcal{M}(s, \mathbf{P})) ds$, the optimal perturbation $\Delta \mathbf{P}$ that minimizes the first-order approximation of $L[\mathcal{M}(s, \mathbf{P} + \Delta \mathbf{P})]$ subject to a constraint on the total squared distance $\int_0^1 |\mathcal{M}(s, \mathbf{P} + \Delta \mathbf{P}) - \mathcal{M}(s, \mathbf{P})|^2 \: \mathrm{d}s = C_0$ is given by:}
\begin{equation}
\Delta \mathbf{P} = -\frac{1}{2\lambda} \left[\int_0^1 \mathbf{M}(s) \: \mathrm{d}s\right]^{-1}\int_0^1 \mathbf{g}(s) \: \mathrm{d}s
\end{equation}
\textit{where $\mathbf{M}(s) = \left(\frac{\partial\mathcal{M}}{\partial \mathbf{P}}\right)^T\frac{\partial\mathcal{M}}{\partial \mathbf{P}}$ is the local metric tensor, $\mathbf{g}(s) = \left(\frac{\partial\mathcal{M}}{\partial \mathbf{P}}\right)^T\nabla \ell(\mathcal{M}(s))$ is the local gradient, and $\lambda$ is a Lagrange multiplier that controls the step size.}

\subsection{Proof of Theorem 1}

We begin by establishing the linearization of the loss functional around the current manifold $\mathcal{M}(s, \mathbf{P})$. For a small perturbation $\Delta \mathbf{P}$, the manifold changes approximately as:

\begin{align}
    \mathcal{M}(s, \mathbf{P} + \Delta \mathbf{P}) \approx \mathcal{M}(s, \mathbf{P}) + \frac{\partial \mathcal{M}}{\partial \mathbf{P}}\Delta \mathbf{P}
\end{align}

Using this linearization, we can approximate the loss functional:

\begin{align}
    L[\mathcal{M}(s, \mathbf{P} + \Delta \mathbf{P})] &\approx \int_0^1 \ell\left(\mathcal{M}(s, \mathbf{P}) + \frac{\partial \mathcal{M}}{\partial \mathbf{P}}\Delta \mathbf{P}\right) \: \mathrm{d}s \\
    &\approx \int_0^1 \left[\ell(\mathcal{M}(s, \mathbf{P})) + \nabla \ell(\mathcal{M}(s, \mathbf{P})) \cdot \frac{\partial \mathcal{M}}{\partial \mathbf{P}}\Delta \mathbf{P}\right] \: \mathrm{d}s + \mathcal{O}(\|\Delta \mathbf{P}\|^2) \\
    &= L[\mathcal{M}] + \int_0^1 \nabla \ell(\mathcal{M}(s, \mathbf{P})) \cdot \frac{\partial \mathcal{M}}{\partial \mathbf{P}}\Delta \mathbf{P} \: \mathrm{d}s + \mathcal{O}(\|\Delta \mathbf{P}\|^2)
\end{align}

Here, we've used a first-order Taylor expansion of $\ell$ around $\mathcal{M}(s, \mathbf{P})$. 

Now, let's consider the distance constraint. The squared distance between the original and perturbed manifolds is:

\begin{align}
    d^2(\mathcal{M}(s, \mathbf{P} + \Delta \mathbf{P}), \mathcal{M}(s, \mathbf{P})) &= \int_0^1 |\mathcal{M}(s, \mathbf{P} + \Delta \mathbf{P}) - \mathcal{M}(s, \mathbf{P})|^2 \: \mathrm{d}s \\
    &\approx \int_0^1 \left|\frac{\partial \mathcal{M}}{\partial \mathbf{P}}\Delta \mathbf{P}\right|^2 \: \mathrm{d}s + \mathcal{O}(\|\Delta \mathbf{P}\|^3) \\
    &= \int_0^1 \Delta \mathbf{P}^T\left(\frac{\partial \mathcal{M}}{\partial \mathbf{P}}\right)^T\left(\frac{\partial \mathcal{M}}{\partial \mathbf{P}}\right)\Delta \mathbf{P} \: \mathrm{d}s + \mathcal{O}(\|\Delta \mathbf{P}\|^3)
\end{align}

For notational convenience, let's define:

\begin{align}
    \mathbf{M}(s) &= \left(\frac{\partial\mathcal{M}}{\partial \mathbf{P}}\right)^T\frac{\partial\mathcal{M}}{\partial \mathbf{P}} &\text{(local metric tensor)} \\
    \mathbf{g}(s) &= \left(\frac{\partial\mathcal{M}}{\partial \mathbf{P}}\right)^T\nabla \ell(\mathcal{M}(s, \mathbf{P})) &\text{(local gradient)}
\end{align}

Our optimization problem now becomes:

\begin{equation}
    \begin{aligned}
    \text{minimize} \quad & \int_0^1 \mathbf{g}(s)^T\Delta \mathbf{P} \: \mathrm{d}s \\
    \text{subject to} \quad & \int_0^1 \Delta \mathbf{P}^T \mathbf{M}(s)\Delta \mathbf{P} \: \mathrm{d}s = C_0
    \end{aligned}
\end{equation}

To solve this constrained optimization problem, we form the Lagrangian:

\begin{equation}
    \mathcal{L}(\Delta \mathbf{P}, \lambda) = \int_0^1 \mathbf{g}(s)^T\Delta \mathbf{P} \: \mathrm{d}s + \lambda\left(\int_0^1 \Delta \mathbf{P}^T \mathbf{M}(s)\Delta \mathbf{P} \: \mathrm{d}s - C_0\right)
\end{equation}

Taking the functional derivative with respect to $\Delta \mathbf{P}$ and setting it to zero:

\begin{align}
    \frac{\delta \mathcal{L}}{\delta(\Delta \mathbf{P})} &= \int_0^1 \mathbf{g}(s) \: \mathrm{d}s + \lambda\int_0^1 2 \mathbf{M}(s)\Delta \mathbf{P} \: \mathrm{d}s = 0\\
    \Rightarrow \int_0^1 \mathbf{g}(s) \: \mathrm{d}s &= -2\lambda\int_0^1 \mathbf{M}(s)\Delta \mathbf{P} \: \mathrm{d}s \\
    \Rightarrow \int_0^1 \mathbf{M}(s)\Delta \mathbf{P} \: \mathrm{d}s &= -\frac{1}{2\lambda}\int_0^1 \mathbf{g}(s) \: \mathrm{d}s
\end{align}

Now, we need to solve for $\Delta \mathbf{P}$. Since $\Delta \mathbf{P}$ is independent of the conditioning variable $s$, we can pull it outside the integral:

\begin{align}
    \int_0^1 \mathbf{M}(s)\Delta \mathbf{P} \: \mathrm{d}s &= \left(\int_0^1 \mathbf{M}(s) \: \mathrm{d}s\right)\Delta \mathbf{P}\\
    \Rightarrow \left(\int_0^1 \mathbf{M}(s) \: \mathrm{d}s\right)\Delta \mathbf{P} &= -\frac{1}{2\lambda}\int_0^1 \mathbf{g}(s) \: \mathrm{d}s\\
    \Rightarrow \Delta \mathbf{P} &= -\frac{1}{2\lambda}\left(\int_0^1 \mathbf{M}(s)\: \mathrm{d}s\right)^{-1}\int_0^1 \mathbf{g}(s) \: \mathrm{d}s
\end{align}

The parameter $\lambda$ controls the step size and can be set to satisfy the distance constraint. In practice, it serves a similar role to the learning rate in gradient descent.

\section{Metric tensors for common parameterizations}

\subsection{Elliptical Manifold}

The manifold is given by:

$$
M(s, \mathbf{P}) = \mathbf{c} + \mathbf{a} \cos(2 \pi s) + \mathbf{b} \sin(2 \pi s)
$$

where $\mathbf{P} = (\mathbf{c}, \mathbf{a}, \mathbf{b})$.

The Jacobian with respect to the parameters $\mathbf{c}$, $\mathbf{a}$, and $\mathbf{b}$ is:

$$
\frac{\partial \mathcal{M}}{\partial \mathbf{P}} = [\mathbf{I}, \cos(2 \pi s) \mathbf{I}, \sin(2 \pi s) \mathbf{I}]
$$

The metric tensor is:

$$
\mathbf{M}(s) = \left(\frac{\partial \mathcal{M}}{\partial \mathbf{P}}\right)^T \frac{\partial \mathcal{M}}{\partial \mathbf{P}} = 
\begin{bmatrix} 
\mathbf{I} & \cos(2\pi s) \mathbf{I} & \sin(2\pi s) \mathbf{I} \\
\cos(2\pi s) \mathbf{I} & \cos^2(2\pi s) \mathbf{I} & \cos(2\pi s) \sin(2\pi s) \mathbf{I} \\
\sin(2\pi s) \mathbf{I} & \cos(2\pi s) \sin(2\pi s) \mathbf{I} & \sin^2(2\pi s) \mathbf{I}
\end{bmatrix}
$$

Integrating the metric tensor over $s \in [0, 1]$ yields \footnote{a large language model was used to assist with evaluating this integral}:

$$
\int_0^1 \mathbf{M}(s) \: \mathrm{d}s = 
\begin{bmatrix} 
\mathbf{I} & 0 & 0 \\
0 & \frac{1}{2} \mathbf{I} & 0 \\
0 & 0 & \frac{1}{2} \mathbf{I} 
\end{bmatrix}
$$

The inverse of the integrated metric tensor is:

$$
\left(\int_0^1 \mathbf{M}(s) \: \mathrm{d}s\right)^{-1} = 
\begin{bmatrix} 
\mathbf{I} & 0 & 0 \\
0 & 2 \mathbf{I} & 0 \\
0 & 0 & 2 \mathbf{I} 
\end{bmatrix}
$$

Given the local gradient $\mathbf{g}(s) = \left(\frac{\partial \mathcal{M}}{\partial \mathbf{P}}\right)^T \nabla \ell(\mathcal{M}(s))$, the optimal update is:

$$
\Delta \mathbf{P} = -\frac{1}{2\lambda} \left(\int_0^1 \mathbf{M}(s) \: \mathrm{d}s\right)^{-1} \int_0^1 \mathbf{g}(s) \: \mathrm{d}s
$$

Substituting the inverse of the integrated metric tensor:

$$
\Delta \mathbf{P} = -\frac{1}{2\lambda} 
\begin{bmatrix} 
\mathbf{I} & 0 & 0 \\
0 & 2 \mathbf{I} & 0 \\
0 & 0 & 2 \mathbf{I} 
\end{bmatrix}
\begin{bmatrix} 
\int_0^1 \nabla \ell(\mathcal{M}(s)) \: \mathrm{d}s \\
\int_0^1 \cos(2 \pi s) \nabla \ell(\mathcal{M}(s)) \: \mathrm{d}s \\
\int_0^1 \sin(2 \pi s) \nabla \ell(\mathcal{M}(s)) \: \mathrm{d}s 
\end{bmatrix}
$$

The final update becomes:

$$
\Delta \mathbf{P} = -\frac{1}{2\lambda} 
\begin{bmatrix} 
\int_0^1 \nabla \ell(\mathcal{M}(s)) \: \mathrm{d}s \\
2 \int_0^1 \cos(2 \pi s) \nabla \ell(\mathcal{M}(s)) \: \mathrm{d}s \\
2 \int_0^1 \sin(2 \pi s) \nabla \ell(\mathcal{M}(s)) \: \mathrm{d}s 
\end{bmatrix}
$$

\subsection{B-Spline Parametrization}

B-splines provide a flexible and numerically stable way to represent curves in weight space. A B-spline manifold is parametrized as:

\begin{equation}
    \mathcal{M}(s, \mathbf{P}) = \sum_{i=0}^n \mathbf{P}_i B_i(s)
\end{equation}

where $\mathbf{P} = (\mathbf{P}_0, \mathbf{P}_1, \ldots, \mathbf{P}_n)$ are the control points in weight space, and $B_i(s)$ are the B-spline basis functions of degree $k$.

Below, we will derive the metric tensor for the parameterization, assuming that the basis points are distributed strictly uniformly over the line. This assumption simplifies the metric, but would require maintaining uniformity through optimization via a constraint.

\subsubsection{Metric Tensor Derivation}

To apply our manifold optimization approach, we need to compute the integrated metric tensor $\mt = \int_0^1 \mathbf{M}(s)ds$. The Jacobian matrix is:

\begin{equation}
    \frac{\partial \mathcal{M}}{\partial \mathbf{P}} = \begin{pmatrix} B_0(s) \mathbf{I} & B_1(s)\mathbf{I} & \cdots & B_n(s)\mathbf{I} \end{pmatrix}
\end{equation}

where $\mathbf{I}$ is the identity matrix with the same dimension as the network parameters. The local metric tensor is:

\begin{equation}
    \mathbf{M}(s) = \left(\frac{\partial\mathcal{M}}{\partial \mathbf{P}}\right)^T\frac{\partial\mathcal{M}}{\partial \mathbf{P}} = 
    \begin{pmatrix} 
        B_0(s)^2 \mathbf{I} & B_0(s)B_1(s)\mathbf{I} & \cdots & B_0(s)B_n(s)\mathbf{I} \\
        B_1(s)B_0(s)\mathbf{I} & B_1(s)^2 \mathbf{I} & \cdots & B_1(s)B_n(s)\mathbf{I} \\
        \vdots & \vdots & \ddots & \vdots \\
        B_n(s)B_0(s)\mathbf{I} & B_n(s)B_1(s)\mathbf{I} & \cdots & B_n(s)^2 \mathbf{I}
    \end{pmatrix}
\end{equation}

To compute the integrated metric tensor, we need to evaluate:

\begin{equation}
    \mt_{ij} = \int_0^1 B_i(s)B_j(s) \: \mathrm{d}s \cdot \mathbf{I}
\end{equation}

\subsubsection{Cubic B-Splines on Uniform Knots}

For cubic B-splines ($k=3$) on a uniform knot sequence with spacing $h$, each basis function $B_i(s)$ is nonzero only over four adjacent knot spans. Note that the spacing can be assumed to be uniform in $s$ such that $h=1/n$ for $n$ knots. 

\subsubsection{Definition of Cubic B-Spline Basis Functions}

For a uniform knot sequence, the cubic B-spline basis function $B_i(s)$ can be explicitly defined as:

\begin{equation}
B_i(s) = \frac{1}{6h^3} \begin{cases}
(s - s_{i-2})^3, & s \in [s_{i-2}, s_{i-1}) \\
(s - s_{i-2})^3 - 4(s - s_{i-1})^3, & s \in [s_{i-1}, s_{i}) \\
(s_{i+2} - s)^3 - 4(s_{i+1} - s)^3, & s \in [s_{i}, s_{i+1}) \\
(s_{i+2} - s)^3, & s \in [s_{i+1}, s_{i+2}) \\
0, & \text{otherwise}
\end{cases}
\end{equation}

A key property is that $B_i(s)B_j(s) = 0$ if $|i-j| > 3$, meaning the metric tensor has a banded structure with bandwidth 3.

\subsubsection{Integration of B-Spline Products}

To evaluate the integrated metric tensor, we need to compute the integrals of products of B-spline basis functions. Due to the compact support and piecewise polynomial nature of B-splines, these integrals can be computed analytically.

The analytical integration of products of B-spline basis functions yields \footnote{a large language model was used to assist with evaluating this integral}:

\begin{equation}
\mt_{ij} = \int_0^1 B_i(s)B_j(s)ds \cdot \mathbf{I} = 
\begin{cases}
\frac{1}{140} \cdot \mathbf{I}, & \text{if } |i-j| = 3 \\
\frac{1}{60} \cdot \mathbf{I}, & \text{if } |i-j| = 2 \\
\frac{11}{140} \cdot \mathbf{I}, & \text{if } |i-j| = 1 \\
\frac{1}{20} \cdot \mathbf{I}, & \text{if } |i-j| = 0
\end{cases}
\end{equation}

For a system with $n+1$ control points (from $\mathbf{P}_0$ to $\mathbf{P}_n$), the integrated metric tensor $\mt$ has the following banded structure:

\begin{equation}
\mt = \frac{1}{420} \cdot 
\begin{pmatrix} 
21 & 33 & 3 & 1 & 0 & \ldots & 0 \\
33 & 21 & 33 & 3 & 1 & \ldots & 0 \\
3 & 33 & 21 & 33 & 3 & \ldots & 0 \\
1 & 3 & 33 & 21 & 33 & \ldots & 0 \\
0 & 1 & 3 & 33 & 21 & \ldots & 0 \\
\vdots & \vdots & \vdots & \vdots & \vdots & \ddots & \vdots \\
0 & 0 & \ldots & 1 & 3 & 33 & 21
\end{pmatrix} \cdot \mathbf{I}
\end{equation}

The inverse of this metric tensor is required for the optimal manifold perturbation as shown in Theorem 1. This matrix is invertible as it is a Gram matrix; this inverse can be precomputed and used for all updates.

\section{Supplement: Algorithmic details}

Below is an algorithmic specification of our update rule.

\begin{algorithm}[H]
\caption{Efficient Manifold Optimization}
\label{alg:manifold_optimization}
\begin{algorithmic}[1]
\REQUIRE Training data $D = \{(\mathbf{x}_i, \mathbf{y}_i, s_i)\}_i^N$, manifold type $\mathcal{M}$ with basis size $n$, network $F$
\STATE Initialize manifold parameters $\mathbf{P}$ for each layer in $F$
\FOR{each epoch}
    \FOR{each batch $\{(\mathbf{x}_i, \mathbf{y}_i, s_i)\}_i^B \subset D$}
        \STATE \texttt{// Forward pass}
        \FOR{each weight $\mathbf{W}(s) = \sum_k^N a_k(s) \mathbf{W}_k$ in each layer in $F$}
            \STATE Compute outputs for basis points matrices $\{\mathbf{W}_k \textbf{x}\}_k^n$
            \STATE For each example, get outputs for the batch as $\{\mathbf{z}_i = \sum_k^n a_k(s) \mathbf{W}_k \mathbf{x}_i\}_i^B$
            \STATE Apply nonlinearities to $\mathbf{z}_i$
        \ENDFOR
        \STATE Compute loss $L = \ell(\hat{\textbf{y}},\textbf{y})$ using final outputs $\hat{\textbf{y}}$
        \STATE \texttt{// Backward pass}
        \STATE Compute gradients $\nabla_\mathbf{P} L$ via auto-differentiation
        \STATE Apply manifold-specific rescaling: $\nabla_\mathbf{P} L \leftarrow \imt \nabla_\mathbf{P} L$
        \STATE Update manifold parameters using rescaled gradients
    \ENDFOR
\ENDFOR
\end{algorithmic}
\end{algorithm}

\section{Supplement: Additional methods details}

Throughout the manuscript, the term `CNN' specifically refers to a network with 3 convolutional layers with [32, 64, 128] filters and kernel size of 3, followed by a MLP with one layer of 512 features. Each convolutional layer is followed by max pooling with a window shape and stride of 2, and all nonlinearities are ReLU.

All experiments in the main manuscript use SGD with a learning rate of 0.01 and momentum of 0.9.

For the `concat' conditioning strategy, the conditioning information was injected into the CNN after the convolutional layers, and before the fully connected layers. For the `embed' strategy, an embedding layer with width 32 was created which sees the conditioning information, and the embedding was concatenated the flattened output of the convolutional filters.

All experments were carried out on Nvidia H100 cards. For consistency, we report the mean and standard deviations of 20 random initialization seeds. For efficiency, these 20 networks were v-mapped in Jax and thus see the same data in the same order, i.e. share a data seed.

\section{Additional experiments}

\begin{figure}[!ht]
    \centering
    \includegraphics[width=0.99\textwidth]{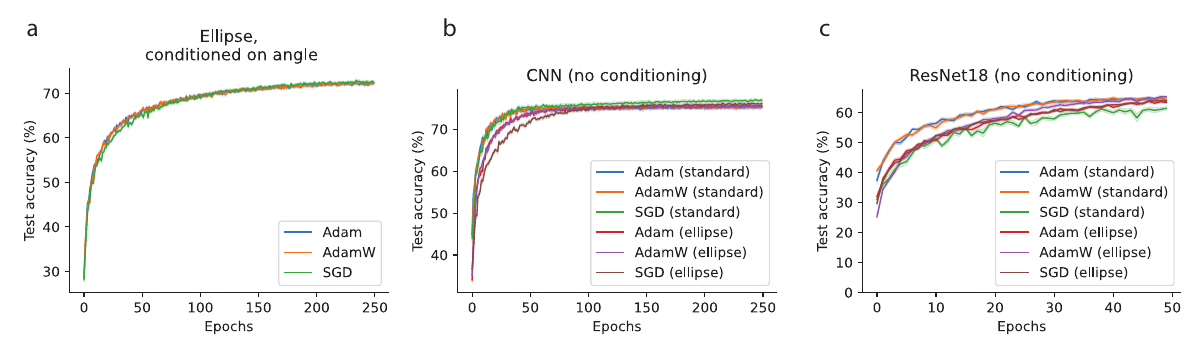}
    \caption{\textbf{a)} We train an elliptical manifold in the space of weights of the same CNN architecture in the main manuscript on rotated CIFAR-10, conditioning on rotation angle by mapping it to ellipse phase. Interestingly, we find that Adam and AdamW do not show meaningful improvements over SGD with momentum. \textbf{b)} Manifolds can also be trained without conditioning on task variables. Here, we train an ellipse on CIFAR-10 using several optimizers, randomizing for each example which network on the manifold is chosen. Convergence accuracy is identical to training a point network. \textbf{c)} Here, we train on the identical task in panel \textbf{b} but using a ResNet18 architecture with LayerNorm. Manifolds and single points (i.e. standard training) perform similarly.}
    \label{fig:supp-train}
\end{figure}

Here, we extend the experiments in the main manuscript to other optimizers, architectures, and to cover the case when the entire manifold is trained on the same objective with the same data rather than conditioned on side information.
The training details are as follows. 

\paragraph{Panel a:} Here, we train an elliptical manifold of CNN weights using Adam and AdamW. As with the SGD, the gradients on the basis points after metric rescaling were fed directly into standard Optax optimizers. Learning rates were tuned in the range 1e-5 to 1e-3 on each optimizer, with optimal rates at 0.01 for SGD and 0.0002 for Adam variants. 

\paragraph{Panel b:} Here, we train an identical CNN architecture with standard optimizers on the standard (not-rotated) CIFAR-10 task, and contrast this to training an elliptical architecture over weights but without conditioning. To ensure that the entire manifold is good at CIFAR-10, a random point on the manifold was used on each example. 

\paragraph{Panel c:} Similar to b, but using a ResNet-18 architecture with LayerNorm normalization layers.

\end{document}